\documentclass[a4paper]{article}

\usepackage{graphicx,amsmath,epsfig,amssymb,bm,multirow,multicol}
\usepackage{algorithm,tabularx,adjustbox}
\usepackage{caption}
\usepackage{subcaption}
\usepackage[noend]{algpseudocode}
\usepackage[super]{nth}
\usepackage{microtype}

\usepackage{INTERSPEECH2022}

\title{Improving Generalization of Deep Neural Network Acoustic Models with Length Perturbation and N-best Based Label Smoothing}
\name{Xiaodong Cui$^1$, George Saon$^1$, Tohru Nagano$^2$, Masayuki Suzuki$^2$, Takashi Fukuda$^2$, Brian Kingsbury$^1$, Gakuto Kurata$^2$}
\address{
  $^1$IBM Research AI, \ IBM T. J. Watson Research Center, \  USA\\
  $^2$IBM Research Tokyo,  \ Japan}
\email{\{cuix,gsaon,bedk\}@us.ibm.com, \{tohru3,szuk,fukuda1,gakuto\}@jp.ibm.com}

\begin{document}

\maketitle
\begin{abstract}
We introduce two techniques, length perturbation and n-best based label smoothing, to improve generalization of deep neural network (DNN) acoustic models for automatic speech recognition (ASR). Length perturbation is a data augmentation algorithm that randomly drops and inserts frames of an utterance to alter the length of the speech feature sequence. N-best based label smoothing randomly injects noise to ground truth labels during training in order to avoid overfitting, where the noisy labels are generated from n-best hypotheses. We evaluate these two techniques extensively on the 300-hour Switchboard (SWB300) dataset and an in-house 500-hour Japanese (JPN500) dataset using recurrent neural network transducer (RNNT) acoustic models for ASR. We show that both techniques improve the generalization of RNNT models individually and they can also be complementary. In particular, they yield good improvements over a strong SWB300 baseline and give state-of-art performance on SWB300 using RNNT models.
\end{abstract}

\noindent\textbf{Index Terms}: length perturbation, label smoothing, n-best hypotheses, deep neural networks, automatic speech recognition

\section{Introduction}
\label{sec:intro}

Generalization is a fundamental problem in machine learning. In ASR, acoustic models with deep neural network (DNN) architectures may suffer from overfitting due to their huge number of parameters. To make DNN models generalize well, techniques such as model regularization (e.g. $\ell_{1}$-norm or $\ell_{2}$-norm regularization \cite{Goodfellow_DeepLearning} and dropout \cite{Hinton_Dropout}) and data augmentation \cite{Cui_DataAug}\cite{Ko_SpeedTempo}\cite{Saon_mixup}\cite{Park_SpecAug} have been broadly used in training. In this paper we introduce two techniques to improve generalization of DNN acoustic models in ASR. One is data augmentation based on length perturbation which randomly drops and inserts frames of an utterance to alter the length of the speech feature sequence. The other is label smoothing based on n-best hypotheses.

Length perturbation compresses and stretches the feature sequence of an utterance in the training to provide a perturbed variant of it having a different length. This is conducted by both frame skipping and frame insertion. Frame skipping and frame insertion as separate approaches have been used in the speech community for different purposes. In most cases, frames are skipped in ASR systems, in a fixed or dynamic manner, for reduced processing time in training or decoding \cite{Miao_FrameSkipping}\cite{Saon_RNNT}\cite{Song_FrameSkipping}\cite{Sunder_DropFrame}. In \cite{Jain_SpliceOut}, SpliceOut is proposed to treat frame skipping as a time masking approach to improve generalization of DNN models in various speech recognition and audio classification tasks. In \cite{Sunder_DropFrame}, it is observed that DropFrame, despite being aimed at reducing training time, may also help to improve performance of end-to-end models. Analogously, frame insertion or time stretching is a common perturbation technique for speech and audio signals in the other direction \cite{Prananta_TimeStretch}\cite{Nguyen_DynTimeStretch}\cite{Mignot_DataAugMusic}. From the length perturbation perspective, the current application of either frame skipping or frame insertion tends to perturb the length of an utterance biased towards one direction. The length perturbation approach investigated in this paper consists of perturbation both ways.

Label smoothing, first introduced in \cite{Szegedy_labsmoothing}, aims to improve generalization in machine learning by avoiding overconfidence over labels. Although researchers are still trying to gain insights into the working mechanism of label smoothing \cite{Muller_labsmoothing}\cite{Xu_labsmoothing}, it has been shown to be helpful in a broad variety of machine learning tasks \cite{Zoph_labsmoothing}\cite{Chorowski_labsmoothing}\cite{Vaswani_transformer}\cite{Zeyer_E2E}. In its conventional setting, label smoothing is accomplished by smoothing a one-hot label vector with a uniform distribution across all class labels under the cross-entropy loss function. Since ASR is essentially a sequence to sequence mapping problem and the acoustic model of interest in this work is recurrent neural network transducers (RNNTs)\cite{Graves_RNNT}\cite{Graves_RNNASR} estimated under the maximum likelihood loss function, we approach label smoothing from a sequence perspective. We choose n-best hypotheses as competing ``classes" and use them as ``noisy" labels in the training with probability.

ASR experiments are carried out on the 300-hour Switchboard (SWB300) dataset \cite{Godfrey_SWB}\cite{Cieri_Fisher} which consists of narrowband speech and an in-house 500-hour Japanese (JPN500) dataset which consists of broadband speech to evaluate the two proposed techniques with various configurations. The acoustic models are RNNTs \cite{Saon_RNNT}\cite{Li_RNNT}\cite{He_StreamingRNNT}. We show that both techniques can improve word error rates (WERs) separately on the two datasets. Moreover, the two techniques can be complementary. In particular, when combining length perturbation and n-best label smoothing, we obtain state-of-the-art WERs on the SWB300 dataset using RNNTs.

The remainder of the paper is organized as follows. Section \ref{sec:lenpb} and Section \ref{sec:nbestls} are devoted to the length perturbation and n-best label smoothing, respectively. Experimental results on SWB300 and JPN500 are reported in Section \ref{sec:exp}. Section \ref{sec:sum} concludes the paper with a summary.

\section{Length Perturbation}
\label{sec:lenpb}

\begin{figure}
     \begin{subfigure}[t]{\linewidth}
         \begin{minipage}[c]{0.5\linewidth}
         \resizebox{\width}{0.8cm}{\includegraphics[scale=0.6]{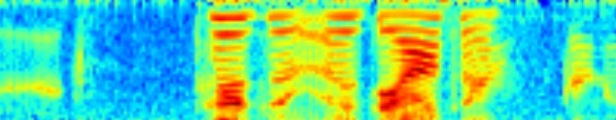}}
         \end{minipage}\hfill
         \begin{minipage}[c]{0.1\linewidth}
         \caption{}\label{fig:lenpb_org}
         \end{minipage}
     \end{subfigure}
     \begin{subfigure}[t]{\linewidth}
         \begin{minipage}[c]{0.5\linewidth}
         \resizebox{\width}{0.8cm}{\includegraphics[scale=0.6]{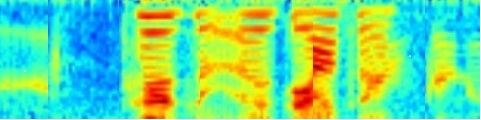}}
         \end{minipage}\hfill
         \begin{minipage}[c]{0.1\linewidth}
         \caption{}\label{fig:lenpb_skip}
         \end{minipage}
     \end{subfigure}
     \begin{subfigure}[t]{\linewidth}
         \begin{minipage}[c]{0.5\linewidth}
         \resizebox{\width}{0.8cm}{\includegraphics[scale=0.6]{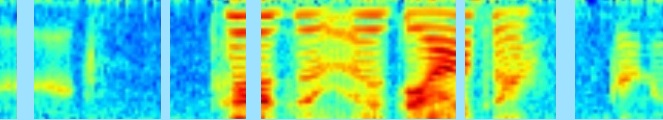}}
         \end{minipage}\hfill
         \begin{minipage}[c]{0.1\linewidth}
         \caption{}\label{fig:lenpb_expand}
         \end{minipage}
     \end{subfigure}
     \begin{subfigure}[t]{\linewidth}
         \begin{minipage}[c]{0.5\linewidth}
         \resizebox{\width}{0.8cm}{\includegraphics[scale=0.6]{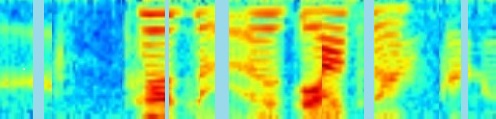}}
         \end{minipage}\hfill
         \begin{minipage}[c]{0.1\linewidth}
         \caption{}\label{fig:lenpb_skip_expand}
         \end{minipage}
     \end{subfigure}\vspace{-0.3cm}
     \caption{Illustration of length perturbation. (a) is the logMel spectrum of an utterance. (b) is the logMel spectrum after frame skipping. (c) is the logMel spectrum after frame insertion. (d) is the logMel spectrum after frame skipping followed by frame insertion.}\label{fig:penpb}\vspace{-0.2cm}
\end{figure}

Implementation of the proposed length perturbation is given in Algorithm \ref{alg:lenpb}. The length of an utterance is perturbed first by skipping frames and then followed by inserting frames, both with a probability. To skip frames, one randomly samples $r_{s}$ percentage of frames from the utterance to operate on. For each sampled frame $x$, $t_{s}$ consecutive frames are dropped starting from $x$ where $t_{s}$ is an integer upper bounded by a hyper-parameter $T_{s}$. Analogously, to insert frames, one randomly samples $r_{p}$ percentage of frames from the utterance to operate on. For each sampled frame $y$, $t_{p}$ consecutive blank frames (zero vectors) are inserted after $y$ where $t_{p}$ is an integer upper bounded by a hyper-parameter $T_{p}$. This is illustrated in Fig.\ref{fig:penpb} where the logMel spectrum of an utterance is demonstrated in Fig.\ref{fig:lenpb_org}. Fig.\ref{fig:lenpb_skip} and Fig.\ref{fig:lenpb_expand} are its perturbed versions by frame skipping and insertion, respectively. Fig.\ref{fig:lenpb_skip_expand} is the overall effect of perturbation after both frame skipping and insertion. All three perturbations are carried out with certain probabilities according to Algorithm \ref{alg:lenpb}.

Length perturbation can help in scenarios where there is a mismatch in the length of utterances between training and test conditions. In addition, by randomly dropping frames, it can also perturb the ``memory" of a sequence model such as long short-term memory (LSTM) \cite{Hochreiter_LSTM} network to avoid simply memorizing the history of the feature sequence in the training. Therefore it can improve the generalization. Furthermore, the applied frame insertion can be viewed as simulating a SpecAug \cite{Park_SpecAug} mechanism for an utterance of longer length (Fig.\ref{fig:lenpb_expand}). It encourages the system to fill in those inserted frames during training, which is also similar in spirit to the filling in frames (FIF) idea in voice conversion \cite{Kaneko_MASKGANFIF}.

\algdef{SE}[SUBALG]{Indent}{EndIndent}{}{\algorithmicend\ }%
\algtext*{Indent}
\algtext*{EndIndent}

\begin{algorithm}[H]
\caption{Length Perturbation of Input Utterances}
\begin{algorithmic}\smallskip
\State $L$ $\leftarrow$ Total number of utterances
\State $p_{s}$ $\leftarrow$ Probability of perturbation by dropping frames;
\State $r_{s}$ $\leftarrow$ Percentage of frames to be perturbed in an utterance;
\State $T_{s}$ $\leftarrow$ Maximum frames to drop;
\State $p_{p}$ $\leftarrow$ Probability of perturbation by inserting frames;
\State $r_{p}$ $\leftarrow$ Percentage of frames to be perturbed in an utterance;
\State $T_{p}$ $\leftarrow$ Maximum frames to insert;

\smallskip

\For  {\ $1 \leftarrow 1, \cdots, L$ \ }
    \State Perturb this utterance by dropping frames with prob. $p_{s}$:
        \Indent
        \State \textbf{DropFrames}($r_{s}$, $T_{s}$)
        \EndIndent
    \State Perturb this utterance by inserting frames with prob. $p_{p}$:
        \Indent
        \State \textbf{InsertFrames}($r_{p}$, $T_{p}$)
        \EndIndent
\EndFor

\smallskip

\Function{\textbf{DropFrames}($r_{s}$, $T_{s}$)}{}
    \State Sample $r_{s}$ percentage of frames from the utterance
    \For {each sampled frame $x$}
        \Indent
        \State Randomly generate an integer $t_{s} \in [1, T_{s}]$
        \State Drop $t_{s}$ consecutive frames starting from $x$
        \EndIndent
    \EndFor
\EndFunction

\smallskip

\Function{\textbf{InsertFrames}($r_{p}$, $T_{p}$)}{}
    \State Sample $r_{p}$ percentage of frames from the utterance
    \For {each sampled frame $y$}
        \Indent
        \State Randomly generate an integer $t_{p} \in [1, T_{p}]$
        \State Insert $t_{p}$ consecutive blank frames after $y$
        \EndIndent
    \EndFor
\EndFunction
\end{algorithmic}\label{alg:lenpb}
\end{algorithm}
\vspace{-0.5cm}

\section{N-best Based Label Smoothing}
\label{sec:nbestls}

Label smoothing introduces a small amount of noise to ground truth labels to avoid training with over-confidence to help generalization. For a classification problem with cross-entropy loss, labels are typically provided as one-hot vectors. Suppose $y$ is a class label for a sample $x$ and there are $K$ classes in total. Label smoothing smoothes the label with a uniform distribution over $K$ classes weighted by $\epsilon$ as shown in Eq.\ref{eqn:ls}
\begin{align}
   \textstyle \tilde{y} = (1-\epsilon) \cdot y + \epsilon \cdot \frac{1}{K}  \label{eqn:ls}
\end{align}
Suppose $p$ is the ground truth (one-hot) distribution, $q$ is the distribution to be learned and $u$ is the uniform distribution. Label smoothing amounts to imposing a regularization term $\sum_{i=1}^{n}H_{i}(u,q)$ to the original cross-entropy term $\sum_{i=1}^{n}H_{i}(p,q)$ as shown in Eq.\ref{eqn:lsxent} where $i=1,\cdots,n$ and $n$ is the total number of samples.
\begin{align}
     \textstyle \mathcal{L} = (1-\epsilon)\sum_{i=1}^{n}H_{i}(p,q) + \epsilon \sum_{i=1}^{n}H_{i}(u,q). \label{eqn:lsxent}
\end{align}

Extending the label smoothing setting in Eq.\ref{eqn:ls} to RNNT training under the maximum likelihood loss is not straightforward as the softmax output of RNNT reflects local decisions while the learning is focused on the whole sequence. From a sequence classification perspective, each sequence may represent a class and all sequences over the output space $\mathcal{Y}$ form a countably infinite set of classes in that sequence space. In this light, we may smooth out the ground truth label sequence with a small number of competing label sequences, which motivates us to investigate the label smoothing strategy based on n-best hypotheses.

Let $\gamma$ be a random variable uniformly distributed in $[0,1]$ and a constant $\epsilon\!\in\![0,1]$. Suppose $\bm{y}$ is a ground truth label sequence and $\Omega_{\bm{y}|\bm{x}} = \{\hat{\bm{y}}_{1},\cdots,\hat{\bm{y}}_{K}\}$ is the n-best set that consists of $K$ n-best hypotheses of $\bm{y}$ given $\bm{x}$. Select uniformly at random from $K$ hypotheses a hypothesis $\hat{\bm{y}}_{i}$, $i \in \{1,\cdots,K\}$, and replace the ground truth label with it with probability $\epsilon$. This is given in Eq.\ref{eqn:nbestls} where $\mathbb{I}(\cdot)$ is the indicator function:
\begin{align}
 \textstyle  \tilde{\bm{y}} = \mathbb{I}(\gamma \leq 1-\epsilon) \cdot \bm{y} + \mathbb{I}(\gamma > 1 - \epsilon) \cdot \hat{\bm{y}}_{i}. \label{eqn:nbestls}
\end{align}

Fig.\ref{fig:nbest} shows an example of a ground truth label sequence and its n-best hypotheses that are used to smooth the ground truth label. The n-best hypotheses are generated by the baseline RNNT models.

\begin{figure}
\footnotesize{
\textbf{this is one this is one of the most highly taxed areas in the country}\\
01 \ \  this is one this is one the most highly taxed areas in the country \\
02 \ \   this is one this is one the most highly tax areas in the country  \\
03 \ \  this is one this is one the most highly taxed areas and country  \\
04 \ \  this one this is one the most highly taxed areas and the country  \\
05 \ \  this is one this is one the most highly tax areas and country  }\vspace{-0.2cm}
\caption{An example of a ground truth label (top line in bold) and its 5 best hypotheses.}\label{fig:nbest}\vspace{-0.5cm}
\end{figure}

\section{Experiments}
\label{sec:exp}


\subsection{SWB300}

The RNNT acoustic model for SWB300 has 6 bi-directional LSTM layers in the transcription network with 1,280 cells in each layer (640 cells in each direction). The prediction network is a single-layer uni-directional LSTM  with 768 cells. The outputs of the transcription network and the prediction network are projected down to a 256-dimensional latent space where they are combined by element-wise multiplication in the joint network. After a hyperbolic tangent nonlinearity followed by a linear transform, it connects to a softmax layer consisting of 46 output units which correspond to 45 characters and the null symbol. The acoustic features are 40-dimensional logMel features extracted every 10ms and their first and second order derivatives. The logMel features are after conversation based mean and variance normalization. Every two adjacent frames are concatenated and appended by a 100-dimensional i-vector \cite{Saon_IVECT} as speaker embedding. Therefore the input to the transcription network is 340 in dimensionality.

The training data goes through three steps of data augmentation. First, it is augmented by speed and tempo perturbation \cite{Ko_SpeedTempo}. This is conducted offline and produces additional 4 replicas of the original training data, which gives rise to about 1,500 hours of training data in total. After the speed and tempo perturbation, mix-up sequence noise is injected where an utterance is artificially corrupted by adding a randomly selected downscaled training utterance from the training set \cite{Saon_mixup}. After that, SpecAugment is applied where the logMel spectrum of a training utterance is randomly masked in blocks in both the time and frequency domains \cite{Park_SpecAug}. Dropout \cite{Hinton_Dropout} is also used in the LSTM layers with a dropout rate of 0.25 and the embedding layer with a dropout rate of 0.05. In addition, DropConnect \cite{Wan_dropconnect} is applied with a rate of 0.25, which randomly zeros out elements of the LSTM hidden-to-hidden transition matrices. A Connectionist Temporal Classification (CTC) \cite{Graves_CTC} model is used to initialize the transcription network.

Optimizer \texttt{AdamW} is used for the training. The learning rate starts at 0.0001 in the first epoch and then linearly scales up to 0.001 in the first 10 epochs. It holds for another 6 epochs before being annealed by $\frac{1}{\sqrt{2}}$ every epoch after the \nth{17} epoch. The training ends after 30 epochs. The batch size is 64 utterances. An alignment-length synchronous decoder \cite{Saon_RNNTdecoding} is used for inference. We measure the WERs with and without an external LM. When decoding with an external LM, density ratio LM fusion \cite{McDermott_LMdenratio} is used. The external LM is trained on a target domain corpus (Fisher and Switchboard) and the source LM is trained only on the training transcripts. The length perturbation is applied before mix-up and SpecAug, all of which are carried out on the fly in the data loader. We evaluate various hyper-parameter configurations for length perturbation ($p_{s}$,$p_{p}$,$r_{s}$,$T_{s}$,$r_{p}$,$T_{p}$) and n-best label smoothing ($K$ and $\epsilon$) on Hub5 2000, Hub5 2001 and RT03 test sets. The data preparation pipeline follows the Kaldi \cite{Povey_Kaldi} s5c recipe.

Table \ref{tab:swb300_lenpb} breaks down the performance of length perturbation on frame insertion (\nth{2}-\nth{5} rows), frame skipping (\nth{6}-\nth{9} rows) and perturbation both ways (\nth{10}-\nth{12} rows), respectively, on the Hub5 2000 test set. The length perturbation is applied in the first 25 epochs and lifted afterwards. From the table, one can observe that both frame insertion and skipping alone can improve WERs. The best WER is obtained when perturbing both ways (avg. 10.7\% without using external LM and 9.4\% when using external LM.)

\begin{table}[tbh]
\caption{Length perturbation using various hyper-parameter configurations on SWB300.}\vspace{-0.3cm}
\begin{adjustbox}{width=\columnwidth,center}
\begin{tabular}{ c | c c c | c c c} \hline
                                           &   \multicolumn{3}{c|}{w/o LM}  &  \multicolumn{3}{c}{w/ LM}   \\ \hline\hline
                                           &    swb    &    ch    &   avg   &   swb   &   ch   &   avg     \\ \hline
   baseline                                &    7.4    &   15.0   &   11.2  &   6.1   &  13.5  &   9.8     \\ \hline\hline
   $p_{p}\!=\!0.6,r_{p}\!=\!0.05,T_{p}\!=\!5$  &    7.1    &   15.1   &   11.1  &   6.0   &  13.3  &   9.7     \\ \hline
   $p_{p}\!=\!0.6,r_{p}\!=\!0.1,T_{p}\!=\!3$   &  \textbf{7.2}    &  \textbf{14.9}   &  \textbf{11.1}  &  \textbf{6.1}   &  \textbf{13.1}  &   \textbf{9.6}    \\ \hline
   $p_{p}\!=\!0.6,r_{p}\!=\!0.1,T_{p}\!=\!7$   &    7.1    &   14.9   &   11.0  &   6.1   &  13.3  &   9.7     \\ \hline
   $p_{p}\!=\!0.7,r_{p}\!=\!0.1,T_{p}\!=\!5$   &    7.3    &   14.7   &   11.0  &   6.1   &  13.3  &   9.7     \\ \hline\hline
   $p_{s}\!=\!0.7,r_{s}\!=\!0.1,T_{s}\!=\!5$   &    7.0    &   14.8   &   10.9  &   6.0   &  13.4  &   9.7     \\ \hline
   $p_{s}\!=\!0.7,r_{s}\!=\!0.1,T_{s}\!=\!7$   &    6.9    &   14.5   &   10.7  &   5.9   &  13.0  &   9.5     \\ \hline
   $p_{s}\!=\!0.8,r_{s}\!=\!0.1,T_{s}\!=\!7$   &   \textbf{6.8}   &  \textbf{14.8}   &  \textbf{10.8}  &  \textbf{5.8}   & \textbf{13.0}  &  \textbf{9.4}     \\ \hline
   $p_{s}\!=\!0.7,r_{s}\!=\!0.1,T_{s}\!=\!9$   &    6.9    &   14.8   &   10.9  &   6.0   &  13.1  &   9.6     \\ \hline\hline
   $p_{s}\!=\!p_{p}\!=\!0.6,r_{s}\!=\!r_{p}\!=\!0.1 $  &  \multirow{2}{*}{7.1}  &  \multirow{2}{*}{14.7} & \multirow{2}{*}{10.9}  &  \multirow{2}{*}{6.1}  &  \multirow{2}{*}{13.2}  &  \multirow{2}{*}{9.7}      \\
   $T_{s}\!=\!7, T_{p}\!=\!3$            &   &   &   &   &    &       \\ \hline
   $p_{s}\!=\!p_{p}\!=\!0.7,r_{s}\!=\!r_{p}\!=\!0.1 $  &  \multirow{2}{*}{\textbf{6.9}}  &  \multirow{2}{*}{\textbf{14.4}} & \multirow{2}{*}{\textbf{10.7}}  &  \multirow{2}{*}{\textbf{5.9}}  &  \multirow{2}{*}{\textbf{12.8}}  &  \multirow{2}{*}{\textbf{9.4}}      \\
   $T_{s}\!=\!7, T_{p}\!=\!3$            &   &   &   &   &    &       \\ \hline
   $p_{s}\!=\!p_{p}\!=\!0.8,r_{s}\!=\!r_{p}\!=\!0.1 $  &  \multirow{2}{*}{7.0}  &  \multirow{2}{*}{14.5} & \multirow{2}{*}{10.8}  &  \multirow{2}{*}{5.9}  &  \multirow{2}{*}{13.0}  &  \multirow{2}{*}{9.5}      \\
   $T_{s}\!=\!7, T_{p}\!=\!3$            &   &   &   &   &    &       \\ \hline
\end{tabular}
\end{adjustbox}
\label{tab:swb300_lenpb}\vspace{-0.2cm}
\end{table}

Table \ref{tab:swb300_nbestls} shows the performance of n-best label smoothing under various $\epsilon$, which controls the probability to replace the ground truth label with a noisy label, and $K$, which is the total number of n-best hypotheses considered. Similarly, the label smoothing is applied in the first 25 epochs and lifted afterwards. The best performance (avg. 10.9\% without using external LM and 9.5\% when using external LM) is achieved when $\epsilon\!=\!0.1$ and $K\!=\!20$.

\begin{table}[tbh]
\caption{N-best based label smoothing using various configurations on SWB300.}\label{tab:swb300_nbestls}\vspace{-0.3cm}
\begin{adjustbox}{width=\columnwidth,center}
\begin{tabular}{ c | c c c | c c c} \hline
                             &   \multicolumn{3}{c|}{w/o LM}  &  \multicolumn{3}{c}{w/ LM}   \\ \hline\hline
                             &    swb    &    ch    &   avg   &   swb   &   ch   &   avg     \\ \hline
   baseline                  &    7.4    &   15.0   &   11.2  &   6.1   &  13.5  &   9.8     \\ \hline
   $\epsilon\!=\!0.1,K\!=\!20$       &   \textbf{7.1}    &   \textbf{14.7}   &   \textbf{10.9}  &  \textbf{6.0}   &  \textbf{13.0}  &  \textbf{9.5}     \\ \hline
   $\epsilon\!=\!0.2,K\!=\!20$       &    7.2    &   14.5   &   10.9  &   6.1   &  13.1  &   9.6     \\ \hline
   $\epsilon\!=\!0.1,K\!=\!30$       &    7.3    &   14.6   &   11.0  &   6.1   &  13.1  &   9.6     \\ \hline
   $\epsilon\!=\!0.2,K\!=\!30$       &    7.1    &   14.9   &   11.0  &   6.0   &  13.0  &   9.5     \\ \hline
\end{tabular}
\end{adjustbox}
\end{table}

Experimental results on combining the two techniques are reported in Table \ref{tab:swb300_lenpb_nbestls} where label smoothing is applied up to 15 epochs and length perturbation is applied between 16 to 30 epochs. After 30 epochs both techniques are lifted and the training continues for another 5 epochs with the learning rates boosted by 2 times. The model used for decoding is after 35 epochs. It shows that the techniques can be complementary. By combining the two techniques we can get avg. WER 10.7\% without using the external LM and 9.2\% with the external LM. This is by far the state-of-the-art single-model result on the Hub5 2000 test set using RNNT, to the best of our knowledge.

\begin{table}[tbh]
\caption{Combination of length perturbation and n-best label smoothing on SWB300.}\label{tab:swb300_lenpb_nbestls}\vspace{-0.3cm}
\begin{adjustbox}{width=\columnwidth,center}
\begin{tabular}{ c | c c c | c c c} \hline
                                           &   \multicolumn{3}{c|}{w/o LM}  &  \multicolumn{3}{c}{w/ LM}   \\ \hline\hline
                                           &    swb    &    ch    &   avg   &   swb   &   ch   &   avg     \\ \hline
   baseline                                &    7.4    &   15.0   &   11.2  &   6.1   &  13.5  &   9.8     \\ \hline
  $\epsilon\!=\!0.1,K\!=\!20,p_{s}\!=\!p_{p}\!=\!0.5$   &  \multirow{2}{*}{6.9}  &  \multirow{2}{*}{15.0} & \multirow{2}{*}{11.0}  &  \multirow{2}{*}{5.8}  &  \multirow{2}{*}{13.0}  &  \multirow{2}{*}{9.4}      \\
  $r_{s}\!=\!r_{p}\!=\!0.1,T_{s}\!=\!5, T_{p}\!=\!5$            &   &   &   &   &    &       \\ \hline
  $\epsilon\!=\!0.1,K\!=\!20,p_{s}\!=\!p_{p}\!=\!0.6$   &  \multirow{2}{*}{7.0}  &  \multirow{2}{*}{14.4} & \multirow{2}{*}{10.7}  &  \multirow{2}{*}{5.9}  &  \multirow{2}{*}{12.7}  &  \multirow{2}{*}{9.3}      \\
  $r_{s}\!=\!r_{p}\!=\!0.1,T_{s}\!=\!5, T_{p}\!=\!5$            &   &   &   &   &    &       \\ \hline
  $\epsilon\!=\!0.1,K\!=\!20,p_{s}\!=\!p_{p}\!=\!0.5$   &  \multirow{2}{*}{\textbf{6.9}}  &  \multirow{2}{*}{\textbf{14.5}} & \multirow{2}{*}{\textbf{10.7}}  &  \multirow{2}{*}{\textbf{5.9}}  &  \multirow{2}{*}{\textbf{12.5}}  &  \multirow{2}{*}{\textbf{9.2}}      \\
  $r_{s}\!=\!r_{p}\!=\!0.1,T_{s}\!=\!7, T_{p}\!=\!3$            &   &   &   &   &    &       \\ \hline
  $\epsilon\!=\!0.1,K\!=\!20,p_{s}\!=\!p_{p}\!=\!0.5$   &  \multirow{2}{*}{6.8}  &  \multirow{2}{*}{14.6} & \multirow{2}{*}{10.7}  &  \multirow{2}{*}{5.9}  &  \multirow{2}{*}{12.7}  &  \multirow{2}{*}{9.3}      \\
  $r_{s}\!=\!r_{p}\!=\!0.1,T_{s}\!=\!7, T_{p}\!=\!3$            &   &   &   &   &    &       \\ \hline
\end{tabular}
\end{adjustbox}
\end{table}

Table \ref{tab:three_testsets} reports the WERs using label smoothing (nbestls), length perturbation (lenpb) and their combination on Hub5 2000, Hub5 2001 and RT03. The external LM is used in the decoding. For comparison, we present the single-model result reported in \cite{Saon_RNNT} as one baseline (\nth{1} row) and the baseline used in this work (\nth{2} row). The difference is the learning rate schedule and the number of epochs. In \cite{Saon_RNNT}, the maximum learning rate is set to 5e-4 and the OneCycleLR policy \cite{Smith_OneCycleLR} is used for 20 epochs. The current baseline gives slightly better performance. The models that give the best performance on Hub5 2000 in Tables \ref{tab:swb300_lenpb}, \ref{tab:swb300_nbestls} and \ref{tab:swb300_lenpb_nbestls}, respectively, are used to evaluate on Hub5 2001 and RT03. As can be seen, although the hyper-parameters of label smoothing and length perturbation are optimized on Hub5 2000, the models generalize well on Hub5 2001 and RT03.

\begin{table}[tbh]
\caption{WERs of length perturbation and n-best label smoothing on Hub5 2000, Hub5 2001 and RT03 test sets.}\vspace{-0.3cm}
\begin{adjustbox}{width=\columnwidth,center}
\begin{tabular}{ l | c c |c c c | c c } \hline
                             &   \multicolumn{2}{c|}{Hub5'00}  &  \multicolumn{3}{c|}{Hub5'01}  &  \multicolumn{2}{c}{RT'03}  \\ \cline{2-8}
                             &    swb    &    ch    &   swb   &   s2p3   &   s2p4  &   swb   &   fsh       \\ \hline
   baseline\cite{Saon_RNNT}  &    6.3    &   13.1   &   7.1   &    9.4   &   13.6  &   15.4  &   9.5       \\ \hline
   baseline                  &    6.1    &   13.5   &   6.7   &    9.6   &   13.4  &   15.7  &   9.0       \\ \hline
   lenpb                     &    5.9    &   12.8   &   6.5   &    9.1   &   13.0  &   15.2  &   8.8       \\ \hline
   nbestls                   &    6.0    &   13.0   &   6.6   &    9.0   &   12.7  &   14.8  &   8.8       \\ \hline
   nbestls+lenpb             &    5.9    &   12.5   &   6.6   &    8.7   &   12.8  &   14.0  &   8.5       \\ \hline
\end{tabular}
\end{adjustbox}
\label{tab:three_testsets}\vspace{-0.5cm}
\end{table}

\subsection{JPN500}

The RNNT acoustic model for JPN500 has 6 bi-directional LSTM layers in the transcription network with 1,280 cells in each layer (640 cells in each direction). The prediction network is a single-layer uni-directional LSTM  with 1,024 cells. The outputs of the transcription network and the prediction network are projected down to a 256-dimensional latent space where they are combined by element-wise multiplication in the joint network. After a hyperbolic tangent nonlinearity followed by a linear transform, it connects to a softmax layer consisting of 3547 output units which correspond to Japanese characters and the null symbol. The acoustic features are 40-dimensional logMel features extracted every 10ms and theirs first and second order derivatives. The logMel features are after utterance based mean normalization. Every four adjacent frames are concatenated. This more aggressive frame skipping is to reduce the length mismatch between the feature sequence and character label sequence. The input to the transcription network is 480 in dimensionality.

There is no data augmentation in the training. But dropout is used in the LSTM layers with a dropout rate of 0.25 and the embedding layer with a dropout rate of 0.05. Optimizer \texttt{AdamW} is used for the training. The learning rate starts at 0.0001 in the first epoch and then linearly scales up to 0.001 in the first 10 epochs. It holds for another 6 epochs before being annealed by $\frac{1}{\sqrt{2}}$ every epoch after the \nth{17} epoch. The model is obtained after 30 epochs. The batch size is 256 utterances. The same alignment-length synchronous decoder is used for inference. No external LM is used in decoding. Both length perturbation and label smoothing are applied in the first 25 epochs and lifted afterwards, respectively. We evaluate various hyper-parameter configurations for length perturbation ($p_{s}$,$p_{p}$,$r_{s}$,$T_{s}$,$r_{p}$,$T_{p}$) and n-best label smoothing ($K$ and $\epsilon$) on 13 real-world test sets from a broad variety of domains and report the average CERs across these test sets.

Table \ref{tab:jap500_lenpb} and Table \ref{tab:jap500_nbestls} show the respective performance of length perturbation and n-best based label smoothing on JPN500 using various hyper-parameter settings. Following a similar trend in SWB300, both techniques help to consistently improve the CERs over the baseline. The length perturbation can reduce the CER from 19.4\% in the baseline to 18.5\% and the label smoothing can reduce to 18.6\%. One interesting observation is that since JPN500 has a more aggressive downsampling strategy (4-frame skipping) in input feature processing compared to that of SWB300 (2-frame skipping) the optimal length perturbation setting for JPN500 tends to favor frame insertion more than frame skipping over SWB300 ($T_{s}\!=\!3,T_{p}\!=\!5$ for JPN500 vs. $T_{s}\!=\!7, T_{p}\!=\!3$ for SWB300). Table \ref{tab:jap500_lenpb_nbestls} shows that the two techniques can also be complementary. In the experiments, the label smoothing is applied up to 15 epochs and length perturbation is applied between 16 to 25 epochs. After 25 epochs both techniques are lifted and the training continues for another 5 epochs. The model used for decoding is after 30 epochs. The combination of the two techniques can achieve a CER of 18.4\%, which amounts to 1\% absolute improvement over the 19.4\% baseline averaging across 13 test sets.

\vspace{-0.1cm}
\begin{table}[tbh]
\caption{Length perturbation using various hyper-parameter configurations on JPN500.}\label{tab:jap500_lenpb}\vspace{-0.3cm}
\centering
\begin{tabular}{ c | c  } \hline
                                                                       &     CER            \\ \hline\hline
  baseline                                                             &     19.4           \\ \hline\hline
  $p_{p}\!=\!0.6,r_{p}\!=\!0.1,T_{p}\!=\!3$                                &     19.0           \\ \hline
  $p_{p}\!=\!0.6,r_{p}\!=\!0.1,T_{p}\!=\!5$                                &     19.2           \\ \hline
  $p_{p}\!=\!0.7,r_{p}\!=\!0.1,T_{p}\!=\!5$                                &    \textbf{18.7}   \\ \hline\hline
  $p_{s}\!=\!0.5,r_{s}\!=\!0.1,T_{s}\!=\!3$                                &    \textbf{18.5}   \\ \hline
  $p_{s}\!=\!0.6,r_{s}\!=\!0.1,T_{s}\!=\!3$                                &     18.7           \\ \hline
  $p_{s}\!=\!0.5,r_{s}\!=\!0.1,T_{s}\!=\!5$                                &     18.7           \\ \hline\hline
  $p_{s}\!=\!p_{p}\!=\!0.6,r_{s}\!=\!r_{p}\!=\!=0.1,T_{s}\!=\!3,T_{p}\!=\!3$         &     19.9           \\ \hline
  $p_{s}\!=\!p_{p}\!=\!0.6,r_{s}\!=\!r_{p}\!=\!=0.1,T_{s}\!=\!3,T_{p}\!=\!5$         &    \textbf{18.6}   \\ \hline
  $p_{s}\!=\!p_{p}\!=\!0.6,r_{s}\!=\!r_{p}\!=\!=0.2,T_{s}\!=\!3,T_{p}\!=\!5$         &    \textbf{18.6}   \\ \hline
\end{tabular}\vspace{-0.2cm}
\end{table}

\vspace{-0.5cm}

\begin{table}[tbh]
\caption{N-best label smoothing using various configurations on JPN500.}\label{tab:jap500_nbestls}\vspace{-0.3cm}
\centering
\begin{tabular}{ c | c  } \hline
                           &     CER            \\ \hline\hline
  baseline                 &     19.4           \\ \hline
  $\epsilon\!=\!0.2,K\!=\!10$      &     19.0           \\ \hline
  $\epsilon\!=\!0.2,K\!=\!20$      &     19.3           \\ \hline
  $\epsilon\!=\!0.2,K\!=\!30$      &  \textbf{18.6}     \\ \hline
  $\epsilon\!=\!0.3,K\!=\!30$      &     18.9           \\ \hline
\end{tabular}
\end{table}

\vspace{-0.5cm}

\begin{table}[tbh]
\caption{Combination of n-best label smoothing and length perturbation on JPN500.}\label{tab:jap500_lenpb_nbestls}\vspace{-0.3cm}
\begin{adjustbox}{width=\columnwidth,center}
\begin{tabular}{ c | c } \hline
                                                                                          &       CER     \\ \hline
   baseline                                                                               &      19.4     \\ \hline
  $\epsilon\!=\!0.2,K\!=\!30,p_{s}\!=\!p_{p}\!=\!0.5, r_{s}\!=\!r_{p}\!=\!0.1,T_{s}\!=\!3, T_{p}\!=\!5$         &      18.5     \\ \hline
  $\epsilon\!=\!0.2,K\!=\!30,p_{s}\!=\!p_{p}\!=\!0.4, r_{s}\!=\!r_{p}\!=\!0.1,T_{s}\!=\!3, T_{p}\!=\!5$         & \textbf{18.4} \\ \hline
  $\epsilon\!=\!0.2,K\!=\!30,p_{s}\!=\!p_{p}\!=\!0.3, r_{s}\!=\!r_{p}\!=\!0.1,T_{s}\!=\!3, T_{p}\!=\!5$         &      18.7     \\ \hline
\end{tabular}
\end{adjustbox}\vspace{-0.5cm}
\end{table}

\subsection{Discussion}

Since the length perturbation perturbs an utterance both ways, it has SpliceOut \cite{Jain_SpliceOut} and DropFrame \cite{Sunder_DropFrame} as a special case of one-way perturbation, which corresponds to the results in Table \ref{tab:swb300_lenpb} and Table \ref{tab:jap500_lenpb} where only $T_{s}$ is used ($T_{p}\!=\!0$). The length perturbation can be used by itself as shown in the JPN500 case or used together with other data augmentation techniques as shown in the SWB300 case. The implementation of the length perturbation may have numerous variations such as the order of frame skipping and insertion or whether frame skipping and insertion should take place in different utterances. This will be further investigated in the future work.

\section{Summary}
\label{sec:sum}

In this paper we introduce length perturbation and n-best based label smoothing to improve the generalization of DNN acoustic modeling. We evaluate the two techniques extensively on both SWB300 and JPN500 datasets and show that both techniques can improve accuracy over strong baselines with RNNT acoustic models. The techniques can be complementary. By combining the two techniques, we have obtained state-of-art single-model results on SWB300 using RNNT.

\bibliographystyle{IEEEtran}

\bibliography{lenpertb_nbestls}

\begin{thebibliography}{10}
\providecommand{\url}[1]{#1}
\csname url@samestyle\endcsname
\providecommand{\newblock}{\relax}
\providecommand{\bibinfo}[2]{#2}
\providecommand{\BIBentrySTDinterwordspacing}{\spaceskip=0pt\relax}
\providecommand{\BIBentryALTinterwordstretchfactor}{4}
\providecommand{\BIBentryALTinterwordspacing}{\spaceskip=\fontdimen2\font plus
\BIBentryALTinterwordstretchfactor\fontdimen3\font minus
  \fontdimen4\font\relax}
\providecommand{\BIBforeignlanguage}[2]{{%
\expandafter\ifx\csname l@#1\endcsname\relax
\typeout{** WARNING: IEEEtran.bst: No hyphenation pattern has been}%
\typeout{** loaded for the language `#1'. Using the pattern for}%
\typeout{** the default language instead.}%
\else
\language=\csname l@#1\endcsname
\fi
#2}}
\providecommand{\BIBdecl}{\relax}
\BIBdecl

\bibitem{Goodfellow_DeepLearning}
I.~Goodfellow, Y.~Bengio, and A.~Courville, \emph{Deep Learning}.\hskip 1em
  plus 0.5em minus 0.4em\relax MIT Press, 2016.

\bibitem{Hinton_Dropout}
N.~Srivastava, G.~Hinton, A.~Krizhevsky, I.~Sutskever, and R.~Salakhutdinov,
  ``Dropout: a simple way to prevent neural networks from overfitting,''
  \emph{Journal of Machine Learning Research}, vol.~15, no.~56, pp. 1929--1958,
  2014.

\bibitem{Cui_DataAug}
X.~Cui, V.~Goel, and B.~Kingsbury, ``Data augmentation for deep neural network
  acoustic modeling,'' \emph{IEEE/ACM Transactions on Audio, Speech, and
  Language Processing}, vol.~23, no.~9, pp. 1469--1477, 2015.

\bibitem{Ko_SpeedTempo}
T.~Ko, V.~Peddinti, D.~Povey, and S.~Khudanpur, ``Audio augmentation for speech
  recognition,'' in \emph{Interspeech}, 2015, pp. 3586--3589.

\bibitem{Saon_mixup}
G.~Saon, Z.~Tuske, K.~Audhkhasi, and B.~Kingsbury, ``Sequence noise injected
  training for end-to-end speech recognition,'' in \emph{International
  Conference on Acoustics, Speech and Signal Processing (ICASSP)}, 2019, pp.
  6261--6265.

\bibitem{Park_SpecAug}
D.~S. Park, W.~Chan, Y.~Zhang, C.-C. Chiu, B.~Zoph, E.~D. Cubuk, and Q.~V. Le,
  ``{SpecAugment}: {A} simple data augmentation method for automatic speech
  recognition,'' in \emph{Interspeech}, 2019, pp. 2613--2617.

\bibitem{Miao_FrameSkipping}
Y.~Miao, J.~Li, Y.~Wang, S.-X. Zhang, and Y.~Gong, ``Simplifying long
  short-term memory acoustic models for fast training and decoding,'' in
  \emph{International Conference on Acoustics, Speech and Signal Processing
  (ICASSP)}, 2016, pp. 2284--2288.

\bibitem{Saon_RNNT}
G.~Saon, Z.~Tueske, D.~Bolanos, and B.~Kingsbury, ``Advancing {RNN} transducer
  technology for speech recognition,'' in \emph{International Conference on
  Acoustics, Speech and Signal Processing (ICASSP)}, 2021.

\bibitem{Song_FrameSkipping}
I.~Song, J.~Chung, T.~Kim, and Y.~Bengio, ``Dynamic frame skipping for fast
  speech recognition in recurrent neural network based acoustic models,'' in
  \emph{International Conference on Acoustics, Speech and Signal Processing
  (ICASSP)}, 2018, pp. 4984--4988.

\bibitem{Sunder_DropFrame}
V.~Sunder, S.~Thomas, H.-K. Kuo, J.~Ganhotra, B.~Kingsbury, and
  E.~Fosler-Lussier, ``Towards end-to-end integration of dialog history for
  improved spoken language understanding,'' in \emph{International Conference
  on Acoustics, Speech and Signal Processing (ICASSP)}, 2022.

\bibitem{Jain_SpliceOut}
A.~Jain, P.~R. Samala, D.~Mittal, and P.~Jyothi, ``{SpliceOut}: a simple and
  efficient audio augmentation method,'' \emph{arXiv preprint
  arXiv:2110.00046}, 2021.

\bibitem{Prananta_TimeStretch}
L.~Prananta, B.~M. Halpern, S.~Feng, and O.~Scharenborg, ``The effectiveness of
  time stretching for enhancing dysarthric speech for improved dysarthric
  speech recognition,'' \emph{arXiv preprint arXiv:2201.04908}, 2022.

\bibitem{Nguyen_DynTimeStretch}
T.-S. Nguyen, S.~Stuker, J.~Niehues, and A.~Waibel, ``Improving
  sequence-to-sequence speech recognition training with on-the-fly data
  augmentation,'' in \emph{International Conference on Acoustics, Speech and
  Signal Processing (ICASSP)}, 2020, pp. 7689--7693.

\bibitem{Mignot_DataAugMusic}
R.~Mignot and G.~Peeters, ``An analysis of the effect of data augmentation
  methods: experiments for a musical genre classification task,''
  \emph{Transactions of the International Society for Music Information
  Retrieval}, vol.~2, no.~1, pp. 97--110, 2019.

\bibitem{Szegedy_labsmoothing}
C.~Szegedy, V.~Vanhoucke, S.~Ioffe, J.~Shlens, and Z.~Wojna, ``Rethinking the
  inception architecture for computer vision,'' in \emph{IEEE conference on
  computer vision and pattern recognition (CVPR)}, 2016, pp. 2818--2826.

\bibitem{Muller_labsmoothing}
R.~Muller, S.~Kornblith, and G.~Hinton, ``When does label smoothing help,'' in
  \emph{Advances in Neural Information Processing Systems (NeuIPS)}, 2019, pp.
  4694--4703.

\bibitem{Xu_labsmoothing}
Y.~Xu, Y.~Xu, Q.~Qian, H.~Li, and R.~Jin, ``Towards understanding label
  smoothing,'' \emph{arXiv preprint arXiv:2006.11653}, 2020.

\bibitem{Zoph_labsmoothing}
B.~Zoph, V.~Vasudevan, J.~Shlens, and Q.~V. Le, ``Learning transferable
  architectures for scalable image recognition,'' in \emph{IEEE conference on
  computer vision and pattern recognition (CVPR)}, 2018, pp. 8697--8710.

\bibitem{Chorowski_labsmoothing}
J.~Chorowski and N.~Jaitly, ``Towards better decoding and language model
  integration in sequence to sequence model,'' in \emph{Interspeech}, 2017, pp.
  523--527.

\bibitem{Vaswani_transformer}
A.~Vaswani, N.~Shazeer, N.~Parmar, J.~Uszkoreit, L.~Jones, A.~N. Gomez,
  L.~Kaiser, and I.~Polosukhit, ``Attention is all you need,'' in
  \emph{Advances in Neural Information Processing Systems (NIPS)}, 2017, pp.
  6000--6010.

\bibitem{Zeyer_E2E}
A.~Zeyer, K.~Irie, R.~Schluter, and H.~Ney, ``Improved training of end-to-end
  attention models for speech recognition,'' in \emph{Interspeech}, 2018, pp.
  7--11.

\bibitem{Graves_RNNT}
A.~Graves, ``Sequence transduction with recurrent neural networks,''
  \emph{arXiv preprint arXiv:1211.3711}, 2012.

\bibitem{Graves_RNNASR}
{A. Graves and A.-r. Mohamed and G. Hinton}, ``Speech recognition with deep
  recurrent neural networks,'' in \emph{International Conference on Acoustics,
  Speech and Signal Processing (ICASSP)}, 2013, pp. 6645--6649.

\bibitem{Godfrey_SWB}
J.~J. Godfrey, E.~C. Holliman, and J.~McDaniel, ``Switchboard: telephone speech
  corpus for research and development,'' in \emph{International Conference on
  Acoustics, Speech and Signal Processing (ICASSP)}, 1992, pp. 517--520.

\bibitem{Cieri_Fisher}
C.~Cieri, D.~Miller, and K.~Walker, ``The {F}isher corpus: a resource for the
  next generation of speech-to-text,'' in \emph{Proceedings of ICLRE}, 2004,
  pp. 69--71.

\bibitem{Li_RNNT}
J.~Li, R.~Zhao, H.~Hu, and Y.~Gong, ``Improving {RNN} transducer modeling for
  end-to-end speech recognition,'' in \emph{Automatic Speech Recognition and
  Understanding Workshop (ASRU)}, 2019.

\bibitem{He_StreamingRNNT}
Y.~He, T.~N. Sainath, R.~Prabhavalkar, I.~McGraw, R.~Alvarez, D.~Zhao,
  D.~Rybach, A.~Kannan, Y.~Wu, R.~Pang, Q.~Liang, D.~Bhatia, S.~Yuan, B.~Li,
  G.~Pundak, K.~C. Sim, T.~Bagby, S.~y.~Chang, K.~Rao, and A.~Gruenstein,
  ``Streaming end-to-end speech recognition for mobile devices,'' in
  \emph{International Conference on Acoustics, Speech and Signal Processing
  (ICASSP)}, 2019, pp. 6381--6385.

\bibitem{Hochreiter_LSTM}
S.~Hochreiter and J.~Schmidhuber, ``Long short-term memory,'' \emph{Neural
  Computation}, vol.~9, no.~8, pp. 1735--1780, 1997.

\bibitem{Kaneko_MASKGANFIF}
T.~Kaneko, H.~Kameoka, K.~Tanaka, and N.~Hojo, ``{MaskCycleGAN-VC}: learning
  non-parallel voice conversion with filling in frames,'' in
  \emph{International Conference on Acoustics, Speech and Signal Processing
  (ICASSP)}, 2021, pp. 5919--5923.

\bibitem{Saon_IVECT}
G.~Saon, H.~Soltau, D.~Nahamoo, and M.~Picheny, ``Speaker adaptation of neural
  network acoustic models using {I}-vectors,'' in \emph{Automatic Speech
  Recognition and Understanding Workshop (ASRU)}, 2013, pp. 55--59.

\bibitem{Wan_dropconnect}
L.~Wan, M.~Zeiler, S.~Zhang, Y.~LeCun, and R.~Fergus, ``Regularization of
  neural networks using {DropConnect},'' in \emph{Proceedings of the 35th
  International Conference on Machine Learning (ICML)}, 2013, pp. 1058--1066.

\bibitem{Graves_CTC}
A.~Graves, S.~Fernandez, F.~Gomez, and J.~Schmidhuber, ``Connectionist temporal
  classification: labelling unsegmented sequence data with recurrent neural
  networks,'' in \emph{Proceedings of the 35th International Conference on
  Machine Learning (ICML)}, 2006, pp. 369--376.

\bibitem{Saon_RNNTdecoding}
G.~Saon, Z.~Tuske, and K.~Audhkhasi, ``Alignment-length synchronous decoding
  for {RNN} transducer,'' in \emph{International Conference on Acoustics,
  Speech and Signal Processing (ICASSP)}, 2020, pp. 7804--7808.

\bibitem{McDermott_LMdenratio}
E.~McDermott, H.~Sak, and E.~Variani, ``A density ratio approach to language
  model fusion in end-to-end automatic speech recognition,'' in \emph{Automatic
  Speech Recognition and Understanding Workshop (ASRU)}, 2019, pp. 434--441.

\bibitem{Povey_Kaldi}
D.~Povey, A.~Ghoshal, G.~Boulianne, L.~Burget, O.~Glembek, N.~Goel,
  M.~Hannemann, P.~Motlicek, Y.~Qian, P.~Schwarz, J.~Silovsky, G.~Stemmer, and
  K.~Vesely, ``The kaldi speech recognition toolkit,'' in \emph{Automatic
  Speech Recognition and Understanding Workshop (ASRU)}, 2011.

\bibitem{Smith_OneCycleLR}
L.~N. Smith and N.~Topin, ``Super-convergence: very fast training of neural
  networks using large learning rates,'' in \emph{Artificial Intelligence and
  Machine Learning for Multi-Domain Operations Applications}, 2019.

\end{thebibliography}

\end{document}